\documentclass[10pt,twocolumn,letterpaper]{article}

\usepackage[pagenumbers]{cvpr} 

\usepackage{graphicx}
\usepackage{amsmath}
\usepackage{amssymb}
\usepackage{booktabs}
\usepackage{enumitem}       

\usepackage[table]{xcolor}  
\usepackage{algorithm}
\usepackage{algorithmic}
\usepackage{bbold}
\usepackage{dblfloatfix}

\usepackage{bbm}
\usepackage{color}
\usepackage{comment}
\usepackage{multibib}
\usepackage{calc}
\usepackage[accsupp]{axessibility}

\def\scalefactor{1.0}
\def\widthfactor{1.0}
\def\figmargin{-9.5pt}

\usepackage[pagebackref,breaklinks,colorlinks]{hyperref}

\newcommand{\revision}[1]{#1}

\usepackage[capitalize]{cleveref}
\crefname{section}{Sec.}{Secs.}
\Crefname{section}{Section}{Sections}
\Crefname{table}{Table}{Tables}
\crefname{table}{Tab.}{Tabs.}

\crefformat{section}{\S#2#1#3} 
\crefformat{subsection}{\S#2#1#3}
\crefformat{subsubsection}{\S#2#1#3}

\usepackage{amsmath,amsfonts,bm}

\def\eqref#1{equation~\ref{#1}}

\def\1{\bm{1}}

\DeclareMathAlphabet{\mathsfit}{\encodingdefault}{\sfdefault}{m}{sl}
\SetMathAlphabet{\mathsfit}{bold}{\encodingdefault}{\sfdefault}{bx}{n}

\def\gD{{\mathcal{D}}}
\def\gE{{\mathcal{E}}}

\def\gR{{\mathcal{R}}}

\DeclareMathOperator*{\argmin}{arg\,min}

\newcommand{\normsq}[1]{\ensuremath{\| #1 \|_2^2}}

\newcommand{\Encinv}{\phi}
\newcommand{\Encsty}{\psi}
\newcommand{\Modsty}{f}
\newcommand{\Dec}{g}
\newcommand{\Head}{h}
\newcommand{\Featsty}{\bm c}
\newcommand{\one}[1]{\mathbbm{1}_{[#1]}}

\newcommand{\authorskip}{\hspace{2.8mm}}

\def\cvprPaperID{1321}
\def\confName{CVPR}
\def\confYear{2022}

\begin{document}

\title{\vspace{-28pt}Towards Robust and Adaptive Motion Forecasting:\\A Causal Representation Perspective\vspace{-6pt}}

\author{
Yuejiang Liu \authorskip Riccardo Cadei\thanks{\vspace{-1pt}Riccardo and Jonas contributed similarly to this work} \authorskip Jonas Schweizer\footnotemark[1] \authorskip Sherwin Bahmani \authorskip Alexandre Alahi\\[1mm]
École Polytechnique Fédérale de Lausanne (EPFL)\\[1mm]
{\tt\small\{firstname.lastname\}@epfl.ch}
\vspace{-4pt}
}

\maketitle

\makeatletter
\renewcommand{\paragraph}{%
  \@startsection{paragraph}{4}%
  {\z@}{1.5ex \@plus 1ex \@minus .2ex}{-1em}%
  {\normalfont\normalsize\bfseries}%
}
\makeatother

\setlength{\abovecaptionskip}{4pt}
\setlength{\belowcaptionskip}{-10pt}

\begin{abstract}
  Learning behavioral patterns from observational data has been a de-facto approach to motion forecasting. Yet, the current paradigm suffers from two shortcomings: \revision{brittle under distribution shifts and inefficient for knowledge transfer}.
In this work, we propose to address these challenges from a causal representation perspective.
We first introduce a causal formalism of motion forecasting, which casts the problem as a dynamic process with three groups of latent variables, namely invariant variables, style confounders, and spurious features.
We then introduce a learning framework that treats each group separately:
(i) unlike the common practice \revision{mixing} datasets collected from different locations, we exploit their subtle distinctions by means of an invariance loss encouraging the model to suppress spurious correlations;
(ii) we devise a modular architecture that factorizes the representations of invariant mechanisms and style confounders to approximate a sparse causal graph;
(iii) we introduce a style contrastive loss that not only enforces the structure of style representations but also serves as a self-supervisory signal for test-time refinement on the fly.
Experiments on synthetic and real datasets show that our proposed method improves the robustness and reusability of learned motion representations, significantly outperforming prior state-of-the-art motion forecasting models for out-of-distribution generalization and low-shot transfer.
  \vspace{-10pt}
\end{abstract}

\section{Introduction}
\label{sec:intro}

Motion forecasting is essential for autonomous systems running in dynamic environments.
Yet, it is a challenging task due to strong spatial-temporal interactions, which arise from two major sources: (i) physical laws (\textit{e.g.}, inertia, goal-directed behaviors) that govern general dynamics; (ii) social norms (\textit{e.g.}, separation distance, left or right-hand traffic) that influence motion styles.
Classical models attempt to describe these interactions based on domain knowledge but often fall short of social awareness in complex scenes \cite{helbing_social_1998, van_den_berg_reciprocal_2011, coscia2018long}.
As an alternative, learning motion representations from observational data has become a \textit{de-facto} approach \cite{alahi_social_2016, salzmann_trajectron_2020, makansi_you_2021}.
In light of rapid progress over the past few years, solving motion forecasting is seemingly just around the corner by pursuing this fashion at larger scales.

However, the promise of the current learning paradigm for motion forecasting is shadowed by two shortcomings:
\begin{itemize}[noitemsep,topsep=0pt,leftmargin=14pt]
    \item struggle to discover physical laws from data, \textit{e.g.}, output inadmissible solutions under spurious shifts \cite{ross_reduction_2011};
    \item inefficient for knowledge transfer, \textit{e.g.}, require a large number of observations to adapt from one environment to another even if the underlying change is sparse \cite{davchev_learning_2021}.
\end{itemize}
These issues do not become any less severe with larger models \cite{sagawa_investigation_2020}. Instead, they are profoundly rooted in the principle of statistical learning that only seeks correlations for the prediction task at hand, regardless of their \emph{robustness} and \emph{reusability} under distribution shifts that may occur in practice {\ifundef{\abbreviated}{
{(illustrated in Figure~\ref{fig:pull})}{}}.

\begin{figure}[t]
  \centering
  \includegraphics[width=\linewidth]{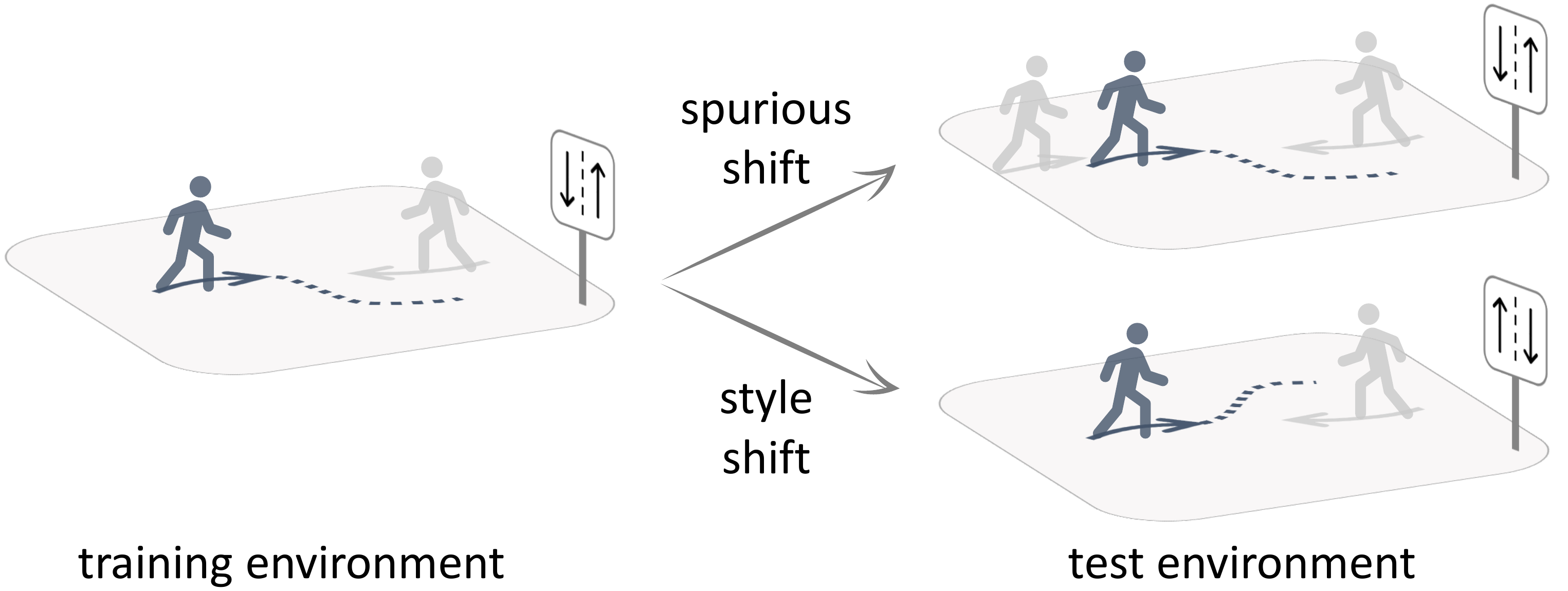}
  \caption{Illustration of motion forecasting under environment changes. \revision{We introduce a framework that enables deep motion representations to robustly generalize to \textit{non-causal shifts} of spurious features, \textit{e.g.}, agent density, and efficiently adapt to new \textit{motion styles}, \textit{e.g.}, from right to left-hand traffic.}}
   \label{fig:pull}
\end{figure}

In this work, we aim to tackle these challenges from a causal representation perspective. Incorporating causal relations into statistical modeling has garnered growing interest lately, as it not only offers a mathematical language to articulate distribution changes \cite{pearl_book_2018, peters_elements_2017} but also brings critical insights to representation learning \cite{scholkopf_toward_2021, wang_desiderata_2021, feder_causal_2021}. Studies in cognitive science have also revealed its paramount importance in the motion context:
few-month-old infants are already able to reason sensibly about physical and social causalities \cite{schlottmann_perceptual_2002}; they can even learn that by sorely observing adult behaviors, without any hands-on experience of their own \cite{waismeyer_learning_2017}. 
How can we build learning algorithms \revision{capable of acquiring such causal knowledge in the same way?}

\ifundef{\abbreviated}{

To this end, we introduce a new formalism of motion forecasting that describes human motion behaviors as a dynamical process with three groups of latent variables: (i) domain-invariant causal variables that account for the physical laws universal to everyone at any place, (ii) domain-specific confounders associated with motion styles, which may vary from site to site, (iii) non-causal spurious features, whose correlations with future motions may change drastically under different conditions.
This causal formalism motivates us to treat each group distinctively with the following three components.

First, we propose to promote causal invariance of the learned motion representations by seeking the commonalities across multiple domains.
Oftentimes, the training dataset is not collected from a single place but comprises multiple subsets from different locations.
Previous work typically merges them into a larger one, \textit{e.g.}, the notable ETH-UCY datasets \cite{pellegrini_improving_2010, lerner_crowds_2007}. 
However, each subset is often inherently different \cite{chen_human_2021}.
Directly combining them not only entails a risk of biases but also destroys the critical information about the stability of correlations.
To address this issue, we train motion forecasting models with a penalty on the variation of empirical risks across environments.
This regularizer encourages the model to suppress spurious features and only exploit causally invariant ones.
As a consequence, the resulting model is close to equally optimal in all environments -- both the ones seen during training and those unseen encountered at test -- for robust generalization.

Second, we design a modular architecture that factorizes the representations of invariant mechanisms and style confounders in a structural way.
One unique property of motion problems is that the style confounder may also vary across environments, but constitute an indispensable part of causal variables for human motions.
To explicitly model their impact, we devise an architecture that contains two encoders responsible for the invariant mechanisms and style confounders separately. 
This modular design approximates the sparse causal graph \cite{obata_neural_1997} in our motion formalism, enabling the model to precisely localize and adapt a small subset of parameters to account for the underlying style shift.

Third, we introduce a style contrastive loss to further strengthen the modular structure of motion styles.
Specifically, we introduce an auxiliary contrastive task that encourages the style encoder to produce an embedding space capturing style relations between different scenes through a simple distance measure.
This peculiar form of discriminative task does not impose prior assumptions on the number of style classes, and is hence particularly suitable for incremental knowledge transfer to new motion styles.
Moreover, when the predicted output is sub-optimal, the style contrastive loss can naturally serve as a self-supervisory signal for test-time refinement on the fly \cite{bau_semantic_2019, liu_collaborative_2020, sun_test-time_2020, li_test-time_2021, liu_ttt_2021}.
By tightly coupling the modular architecture design with the style contrastive loss, our method makes effective use of the knowledge stored in the style encoder during both training and deployment.

We evaluate the proposed method in two settings: synthetic simulation datasets and controlled real-world experiments.
In the presence of spurious correlations, motion forecasting models trained by our invariant loss demonstrate superior out-of-distribution (OOD) generalization ability over previous methods.
Under variations of motion styles, our proposed modular architecture and style loss greatly improve the transferability of forecasting models in the low-shot setting.
We hope our findings will pave the way for a tight integration of causal modeling and representation learning in the motion context, a largely under-explored yet highly promising direction towards reliable and adaptive autonomy.
Our code is available at \url{https://github.com/vita-epfl/causalmotion}.

}
{
To this end, we propose a method that explictly incorporates causal invariance and structure into motion representation learning in order to promote its robustness and transferability under common types of distribution shifts.
}

\begin{figure*}[t]
\vspace{\figmargin}
\centering
\includegraphics[width=14cm]{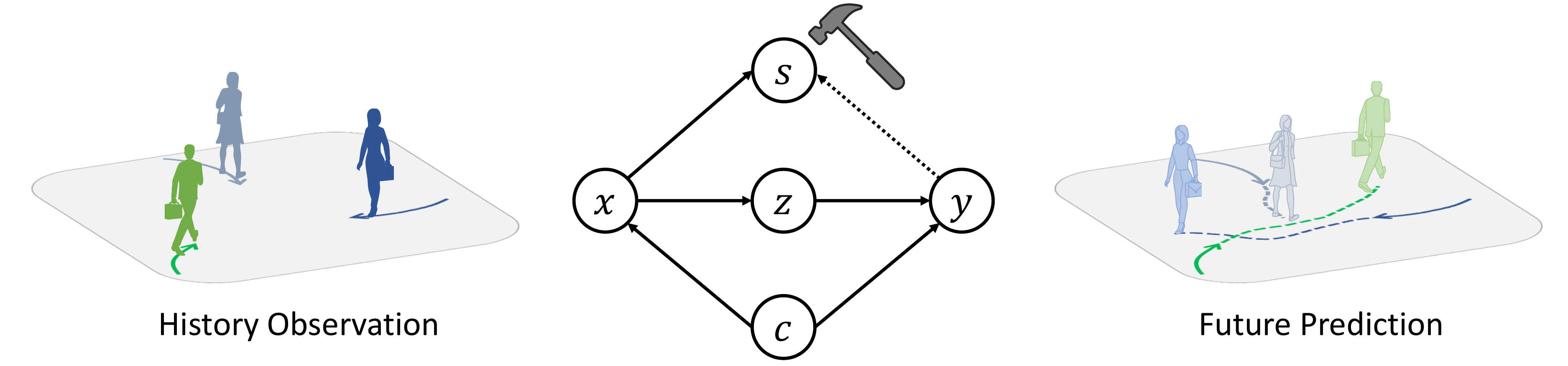}
\caption{Our causal formalism of motion forecasting. We cast the human motion problem as a dynamic process with three groups of latent variables: domain-invariant physical laws ($\bm z$), domain-specific style confounders ($\bm c$), and non-causal spurious features ($\bm s$). 
The spurious features are not parents of future movements ($\bm y$) in the causal graph, \textit{e.g.}, no edges or anti-causal (dotted line), and their statistical correlations may vary drastically under different conditions.
This formalism motivates our design and training of forecasting models to promote the robustness and reusability of the learned motion representation.}
\label{fig:formalism}
\end{figure*}

\section{Related Work}
\label{sec:related}
\paragraph{Motion forecasting.} 
Modern motion forecasting models \cite{alahi_social_2016, gupta_social_2018, huang_stgat_2019, salzmann_trajectron_2020, kothari_human_2021} are largely built with neural networks and trained with the maximum likelihood principle. Despite strong performance for short-range predictions within the training domain, they often struggle to generalize under covariate shift. Recently, a couple of works proposed to promote their robustness using negative data augmentation \cite{liu_social_2021, zhu_motion_2021}. However, designing negative examples of high-dimension, \textit{e.g.}, long sequences, can be difficult in practice. Our work explores a causality-inspired alternative that does not require hand-engineered interventions over training data and is hence more theoretically grounded and algorithmically generic. 

Closely related to ours, another recent work \cite{chen_human_2021} attempts to mitigate biases in motion datasets through counterfactual analysis. Our method differs from theirs in three aspects: (i) their approach learns to estimate dataset biases before subtracting them in the feature space, whereas our method aims to directly suppress biased features; (ii) their approach inherits the merge-and-shuffle convention, which destroys some critical information for bias estimation; in contrast, we keep each subset separately and exploit unstable correlations across environments; (iii) the counterfactual problem formulated in their approach is generally difficult to solve (\textit{cf.} Pearl's ladder of causation \cite{pearl_book_2018}); conversely, we consider spurious correlations from the interventional perspective, which is easier to tackle in practice.

\paragraph{Causal learning.}
The intersection of causal inference and machine learning has been a vibrant area of research in the past few years \cite{peters_elements_2017, wang_desiderata_2021, feder_causal_2021}.
Some earlier works attempted to identify causal structures from observational or interventional data \cite{heinze-deml_causal_2018, vowels_ya_2021}.
Examples include score-based \cite{aragam_2015_concave, scanagatta_2016_learning, huang_2018_generalized}, constraint-based \cite{jaber_2020_causal, huang_2020_causal, yang_2018_char, triantafillou_2015_constraint}, conditional independence test \cite{bellot_cond_2019, shi_double_2020, fukumizu_2007_kernel, zhang_kernel-based_2011}, continuous optimization \cite{zheng_dags_2018, ma_2014_detecting, yang_2021_causalvae} and many others \cite{janzing_2012_information, choi_2020_bayesian, tsamardinos_2006_max}.
While these methods are theoretically appealing, they are often practically restricted to classical problems that assume direct access to high-level causal variables rather than the low-level observations present in modern problems \cite{scholkopf_toward_2021}.

More recently, several different approaches have been proposed to automatically discover causal variables of interest from low-level data.
One notable line of work lies in disentangled representation learning \cite{bengio_representation_2013, chen_infogan:_2016, burgess_understanding_2018, higgins_towards_2018}, which is closely tied to independent causal mechanisms \cite{parascandolo_learning_2018, suter_robustly_2019}.
However, separating independent factors of variation in an unsupervised manner is often exceedingly challenging without strong assumptions \cite{locatello_challenging_2019}.
As an alternative, a few other recent works seek casually invariant representations by exploiting observational data collected under different setups \cite{buhlmann_invariance_2018,arjovsky_invariant_2020,ahuja_invariant_2020,krueger_out--distribution_2021,rosenfeld_risks_2020}.
Our work also falls into this category: we reveal the strengths and weaknesses of the invariant learning principle in the motion context and propose to tightly integrate invariant representation with structural architectural design based on domain knowledge.

\paragraph{Distribution shifts.} 

Previous methods tackle the challenge of distribution shifts from three main paradigms: domain generalization, domain adaptation, and transfer learning. 
Domain generalization is the most ambitious one, which aims to learn models that can directly function well in related but unseen test distributions \cite{blanchard_generalizing_2011,gulrajani_search_2020}.
Recent literature has proposed a variety of solutions, such as distributionally robust optimization \cite{delage_distributionally_2010,duchi_learning_2021,rahimian_distributionally_2019}, adversarial data augmentation \cite{madry_towards_2018,volpi_generalizing_2018}.
Yet, these techniques often rely on strong assumptions on the test distribution, which may not hold in practice.

Domain adaptation is another popular approach that relaxes these assumptions by allowing a learning algorithm to observe a set of unlabelled test samples.
Modern methods of this kind typically attempt to learn an embedding space where the training and test samples are subject to similar feature distributions through divergence minimization \cite{gretton_kernel_2012,long_deep_2017,sun_correlation_2017,zellinger_central_2016} or adversarial training \cite{ganin_domain-adversarial_2016, tzeng_adversarial_2017}. While this approach has been shown effective in a variety of supervised tasks \cite{wang_deep_2018, csurka_deep_2020, zhao_multi-source_2020}, it is not well suited for motion forecasting where labels in the form of future trajectories are fairly easy to acquire without human annotation but sample efficiency matters crucially.

Previous work in the third category -- transfer learning given limited data -- often leverages special architecture designs, \textit{e.g.}, external memory \cite{graves_neural_2014, miller_key-value_2016}, or transfer-oriented objectives, \textit{e.g.}, meta-learning \cite{finn_model-agnostic_2017}.
Some of these techniques have also been applied to motion forecasting \cite{gui_few-shot_2018, zang_few-shot_2020}.
Our work differs from them in that we adopt a causal approach and propose a unified learning framework that facilitates both robust generation and fast adaptation to common types of distribution shifts in motion forecasting.

\section{Method}
\label{sec:method}
\ifundef{\abbreviated}
{

The conventional learning paradigm for motion forecasting only seeks statistical patterns in the collected observational data at hand, regardless of their robustness and reusability under distribution shifts. 
As a result, existing models often struggle to effectively generalize or adapt to new environments.

In this section, we address these challenges by 
(i) formalizing the motion forecasting problem from a causal representation perspective and (ii)
explicitly promoting the causal invariance and structure of the learned motion representations through three algorithmic components. 

}{}

\begin{figure*}[t]
\vspace{\figmargin}
\centering
\includegraphics[width=17.2cm]{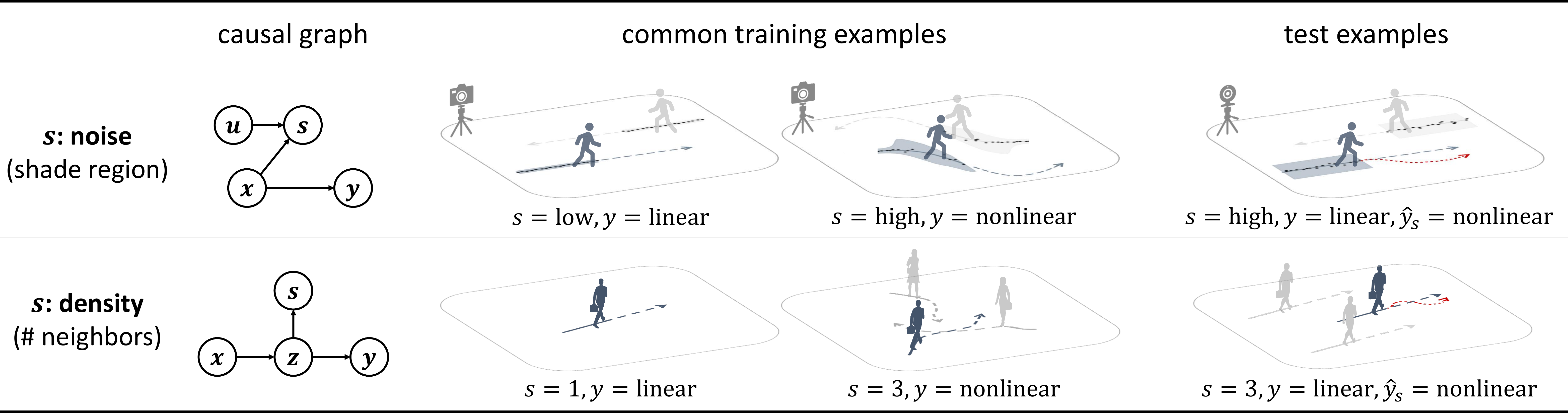}
\caption{Illustration of spurious correlations in motion forecasting. 
The curvature of the target trajectory $\bm y$ is often correlated with spurious features $\bm s$, such as observation noises and agent densities.
Yet, such correlations are not robust.
For instance, the noise level may correlate with $\bm y$ in different ways between training and test, due to the change of exogenous variables $\bm u$, \textit{e.g.}, sensing devices.
Likewise, the agent density and $\bm y$ are not causally related, but confounded by invariant features $\bm z$, \textit{e.g.}, neighboring environment.
In the two illustrated cases, the test example is much closer to the left training example than to the right one in essence.
However, models built upon the spurious correlations in the training examples may output erroneous predictions $\bm \hat y_s$ even on simple test examples.
}
\label{fig:illustration}
\end{figure*}

\subsection{Formalism of Motion Forecasting}

\paragraph{Preliminary.} 

Consider a motion forecasting problem in multi-agent environments. 
For a scene of $M$ agents, let $s_t = \{s_t^1, \cdots, s_t^M\}$ denote their joint state and $s_t^i = (x_t^i, y_t^i)$ denote the state of an individual agent $i$ at time $t$. 
The model takes an input sequence of past observations $\bm x=(s_1, \cdots, s_{t})$ in order to predict their states in the future $\bm y=(s_{t+1}, \cdots, s_{T})$ up to time $T$.
Modern forecasting models are largely built with encoder-decoder neural networks, where the encoder $\Encinv(\cdot)$ extracts a compact motion representation $\bm z$ of the past observations and the decoder $\Dec(\cdot)$ rolls out the predicted trajectory $\bm {\hat y}$.

The training data $\gD$ is often collected from a set of $K$ environments $\gE = \{e_1, e_2, \ldots, e_K\}$. Previous work typically merges them into a large dataset and assumes the mixture as a representative of the unseen test environment $\Tilde{e}$. Under this assumption, the model is trained to minimize the empirical risk:
\begin{equation}
\label{eq:erm}
    \gR(\Encinv,\Dec) := \frac{1}{|\gD|} \sum_{(\bm x,\bm y)\in \gD} \mathcal{L}_{\rm task}(\Dec(\Encinv(\bm x)), \bm y).
\end{equation}
where $\mathcal{L}_{\rm task}$ is the loss function of the motion forecasting task, such as mean square error (MSE) or negative log-likelihood (NLL).
However, the i.i.d. assumption does not always hold in practice. In fact, recent work \cite{chen_human_2021} has shown that the test environment can be significantly different from the training ones in the widely used ETH-UCY benchmark. We will next introduce a causal formalism of motion forecasting that allows us to formulate this challenge and design solutions to address it.
\paragraph{Causal formalism.}

Motion behaviors are essentially dynamic processes governed by latent variables, such as physical laws, traffic rules and social norms.
To build accurate predictive models, the conventional learning paradigm typically aims to discover these latent variables and model their correlations with the observed future states. 
However, the learned correlations may vary across environments and thus fail to generalize at test time.


To tackle this fundamental challenge, we introduce a new formalism of motion forecasting through the lens of causality. As shown in Figure~\ref{fig:formalism}, we categorize the latent variables into three groups: 
\begin{itemize}[noitemsep,topsep=0pt,leftmargin=14pt]
    \item invariant variables: physical laws that are universal to everyone at any place;
    \item hidden confounders: motion styles that may vary from site to site in a local and sparse manner;
    \item spurious features: other variables, \textit{e.g.}, level of noises, that are not direct causes of future motion.
\end{itemize}
Neither the second nor the third group has stable correlations with the target future motion across environments. Yet, they may lead to distinctive effects on forecasting models. Spurious correlations can become drastically different in new settings, resulting in catastrophic errors, as illustrated in Figure~\ref{fig:illustration}. In comparison, variations of motion styles are often more restricted. Models that fail to capture the correct motion style may suffer from inaccurate predictions but should still output plausible solutions subject to physical laws. We will next describe three algorithmic components that treat spurious features and hidden confounders differently in order to promote the robustness and reusability of learned motion representations. 

\subsection{Causal Invariant Forecasting} \label{sec:method_robust}

\paragraph{Invariant principle.} 
By definition, invariant features should have identical joint distributions with the target variable (future motion) across different environments, whereas the non-invariant ones are the opposite. 
This distinction can be formulated as a necessary condition for the domain invariant predictor, \textit{i.e.}, $\Dec \circ \Encinv$ is equally optimal in every environment \cite{peters_causal_2016}. More formally, our goal is to solve the following problem: 
\begin{equation}
\label{eq:irm}
\begin{aligned}
\min_{\Encinv,\Dec} \quad  & \frac{1}{|\gE|}\sum_{e\in\gE} \gR^e(\Encinv,\Dec) \\
\textrm{s.t.} \quad & \Dec \in\argmin_{\Dec^*} \gR^e(\Encinv,\Dec^*)\quad\forall e\in\gE,
\end{aligned}
\end{equation}
where $\Dec^*$ is an optimal predictor built on top of the extracted features in an individual environment $e$. Intuitively, if a learned forecasting model can perform similarly well across multiple training environments, it is more likely to generalize to another related test environment $\Tilde{e}$ as well. 

\begin{figure*}[t]
    \centering
    \vspace{\figmargin}
    \includegraphics[width=0.98\linewidth]{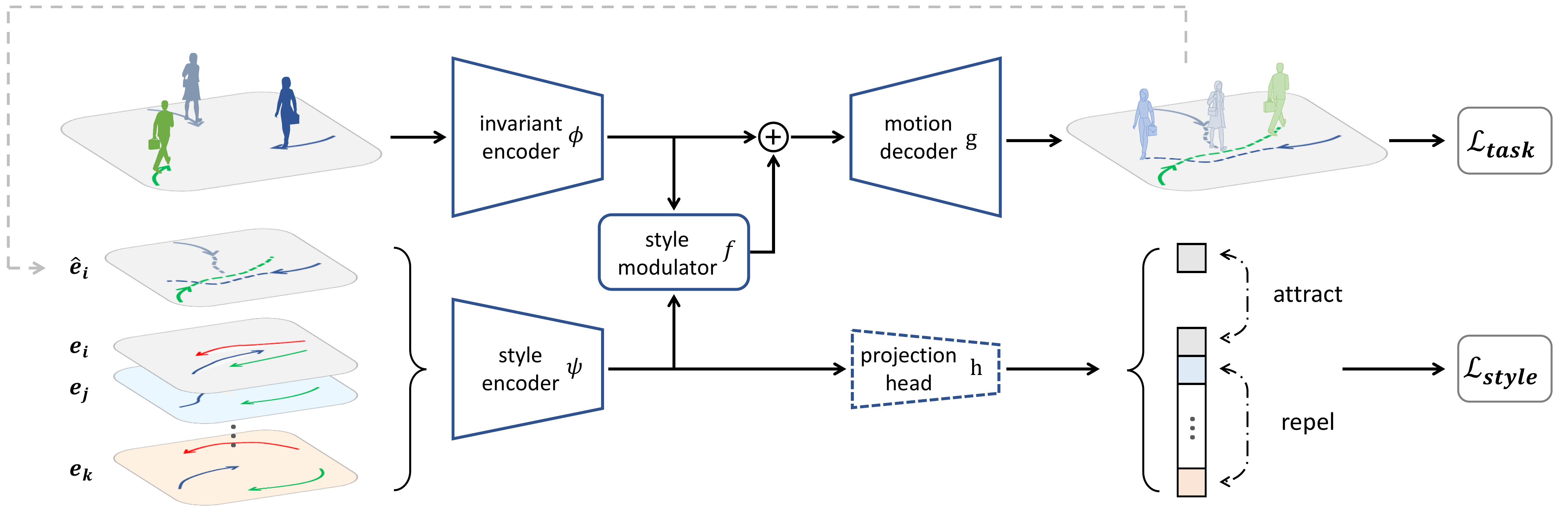}
    \caption{Our modular forecasting model contains two separate encoders for universal laws and style confounders, respectively.
    The model is built in three steps:
    (i) learn an invariant predictor based on the first encoder $\Encinv$, with the goal to be equally optimal in all training environments (\cref{sec:method_robust}),
    (ii) learn an embedding space based on the second encoder $\Encsty$ to capture the style relation between different scenes (\cref{sec:method_style}),
    (iii) incorporate domain-specific style features into the forecasting model by training $\Encsty$, $\Modsty$, $\Dec$ and $\Head$ on the main task and the auxiliary style contrastive task jointly (\cref{sec:method_modular}-\cref{sec:method_style}).
    }
    \label{fig:modular}
\end{figure*}

\paragraph{Invariant loss.} The exact form of the invariant learning principle (Eq.~\ref{eq:irm}), however, leads to a bi-level optimization problem, which is difficult to solve in practice. Recent work \cite{arjovsky_invariant_2020, rosenfeld_risks_2020} proposed to relax it to a gradient norm penalty over the empirical risk $\gR^e$ in each training environment: 
\begin{align}
    \label{eq:irmv1}
    \min_{\Encinv,\Dec}\quad \frac{1}{|\gE|}\sum_{e\in\gE} \left[\gR^e(\Encinv,\Dec) + \lambda\normsq{\nabla_{\Dec}\gR^e(\Encinv,\Dec)}\right].
\end{align}
This objective prevents the forecasting model from learning an \textit{average} effect of spurious features on future trajectories and enforces the model to solve it the hard way by seeking universal mechanisms behind motion behaviors. We will show in Sec.~\ref{sec:exp_spurious} that this technique can greatly improve the robustness of the forecasting model against \revision{distribution shifts of spurious features}. 

However, the strength of suppressing spurious features comes with a clear drawback, \textit{i.e.}, the learned representation tends to erroneously drop the motion styles that change across environments. This may cause inaccurate predictions in both training and test environments.
To cope with this issue, we next introduce a modular architecture that allows the model to properly structure the knowledge and strategically adapt from one style to another. 

\subsection{Modular Forecasting Model} \label{sec:method_modular}

Most recent forecasting models are built with dense connections at their core, albeit with some detailed differences. On the one hand, this design principle is very powerful when the training data is sufficient; on the other hand, it often lands in a highly inter-twined architecture that lacks semantic structure. As such, one may have to update the whole model \revision{even if distribution shifts only arise from variations of motion styles. This fine-tuning convention} inevitably leads to low sample efficiency for transfer learning. Ideally, a forecasting model would preserve a clear structure of the learned knowledge, separate the impacts of physical laws and motion styles on motion behaviors, and approximate the high-level sparse causal graph in Figure~\ref{fig:formalism}.

To achieve this goal, we devise a modular network that consists of two encoders and one decoder. The first encoder $\Encinv$ is trained to compute domain-invariant features, as described in \cref{sec:method_robust}. Subsequently, we introduce a second encoder $\Encsty$ which aims to capture features of motion styles varying across domains.
\revision{Given some style observations $\bm o$ from a particular environment $e$, the role of $\Encsty$ is to produce a latent representation of the style confounders $\Featsty$.
One key difference between the input to the style encoder and that to the invariant encoder lies in that the former is one (or multiple) long sequence where the motion style is fully observable, whereas the latter is the past trajectory $\bm x$, which may not contain sufficient information about the underlying motion style, \textit{e.g.}, before interactions.}
More formally, our modular network predicts the future trajectory as follows:  
\ifundef{\abbreviated}
{
\begin{equation}
\begin{aligned}
    &\bm z = \Encinv(\bm x), \qquad
    \Featsty = \Encsty(\bm o), \\
    &\bm {\tilde z} = f(\bm z,\bm c) + \bm z, \quad
    \bm {\hat y} = g(\bm {\tilde z}),
\end{aligned}
\end{equation}
}
{
\begin{equation}
    \bm z = \Encinv(\bm x), \qquad
    \Featsty = \Encsty(\bm o), \qquad 
    \bm {\tilde z} = f(\bm z,\bm c) + \bm z, \qquad
    \bm {\hat y} = g(\bm {\tilde z}),
\end{equation}
}
where $\bm {\tilde z}$ is the latent feature that incorporates both $\bm z$ and $\bm c$, and the style modulator $\Modsty$ can be modeled by a small multilayer perceptron (MLP).
Here, we can also compute $\Featsty$ based on multiple scene observations from the same environment, \textit{e.g.}, averaging several style feature vectors, to obtain a more robust estimate of the motion style. 
As shown in Figure~\ref{fig:modular}, our modular design allows us to precisely localize and fine-tune a small subset of parameters to account for the underlying style shift.

\subsection{Style contrastive loss} \label{sec:method_style}

Our modular forecasting model composed of multiple sub-networks can be practically difficult to train, especially in the few-shot transfer setting where data collected from the new environment is limited. To overcome this challenge, we introduce a style contrastive loss, which aims to not only strengthen the modular structure of motion representation during training but also allows for reusing the encoded style knowledge at test time. 

\paragraph{Style contrastive learning.}
Ideally, the feature vector produced by the style encoder should not only provide the basic style information for predicting the future motion accurately but also properly capture the style relation between different scenes. We formulate this intuition into an auxiliary task in the form of supervised contrastive learning.
Specifically, we consider two scene observations  from the same environment as a pair of positive samples, whereas those from different environment as negative pairs. We map the style feature $\bm c$ to a projected embedding $\bm p$ by a small head $\Head(\cdot)$. The style contrastive loss for a positive pair of samples $(i,j)$ is as follows, 
\begin{equation}
\label{eq:contrastive}
    \mathcal{L}_{\rm style} = -\log \frac{\exp(\mathrm{sim}( \bm p_i, \bm  p_j)/\tau)}{\sum_k \one{k=j \lor \bm e_k \neq \bm  e_i} \exp(\mathrm{sim}(\bm p_i, \bm p_k)/\tau)},
\end{equation}
where $\one{\bm e_k \neq \bm  e_i}$ is an indicator function equal to $1$ if and only if the two samples $i$ and $k$ are drawn from the same environment, $\tau$ is a temperature parameter and $\mathrm{sim}(\bm u,\bm v) = \bm u^\top \bm v / \lVert\bm u\rVert \lVert\bm v\rVert$ denotes the dot product between normalized $\bm u$ and $\bm v$ (cosine similarity).


One key advantage of the proposed style contrastive loss over the conventional classification loss is that it does not impose any assumptions about the number of domain classes in the design of the projection head $h$. This property allows the model to incrementally bootstrap from the knowledge already learned about the existing styles to some additional ones without changing the shape of $h$ or learning any parameters from scratch. This is particularly beneficial as in the transfer setting where the number of additional styles is not known a priori.

\ifundef{\abbreviated}{

Overall, we train our entire modular forecasting model in three steps:
\begin{enumerate}[noitemsep,topsep=0pt,leftmargin=16pt]
    \item train the predictor backbone $\Encinv $ and $\Dec$ based on the invariant loss (Eq.~\ref{eq:irmv1});
    \item train the style embedding $\Encsty$ and $\Head$ based on the style contrastive loss (Eq.~\ref{eq:contrastive});
    \item train $\Encsty$, $\Modsty$, $\Dec$ and $\Head$ on the task loss (Eq.~\ref{eq:erm}) and style loss (Eq.~\ref{eq:contrastive}) jointly while freezing the invariant encoder $\Encinv$.
\end{enumerate}
In the presence of style shifts, we fine-tune a subset of parameters, \textit{e.g.}, the style modulator $\Modsty$, in order to efficiently adapt the model from the learned domains to a new one.

}{}

\paragraph{Test-time style refinement.}
One common phenomenon in transfer learning is that the model fine-tuned on only a few samples remains sub-optimal in the new environment.
To alleviate this performance gap, we reuse the style contrastive loss as a self-supervisory signal for test-time refinement on the fly.
\revision{Concretely, we feed the predicted output back as an input to the style encoder, examine its style consistency with other observed samples, and iteratively adjust the internal feature $\bm {\tilde z}$.
Here, the variables to optimize are no longer the model weights but rather the feature activations for each test instance.
The refinement process gradually reduces the distances between the predicted output and the reference examples of the same style in the learned embedding space.}
By tightly coupling the modular architecture design with the style contrastive loss, our method enables the effective use of the auxiliary contrastive task during both training and deployment.

\section{Experiments}
\label{sec:experiment}
We evaluate our proposed method on two types of forecasting models (recurrent STGAT \cite{huang_stgat_2019} and feedforward PECNet variant \cite{mangalam_it_2020}) under distribution shifts of spurious features or style confounders. 
In the considered forecasting task, a model processes the past 8 time steps (3.2 seconds) of human trajectories in the scene to then predict their future movements in the following 12 (4.8 seconds) time steps.
Identical to many prior works \cite{alahi_social_2016, gupta_social_2018, salzmann_trajectron_2020}, we evaluate forecasting models on two metrics:
\begin{itemize}[noitemsep,topsep=0pt,leftmargin=14pt]
    \item \textit{Average Displacement Error} (ADE): the average Euclidean distance between the predicted output and the ground truth over all predicted time steps.
    \item \textit{Final Displacement Error} (FDE): the Euclidean distance between the predicted final destination and the true final destination at the end of the prediction horizon.
\end{itemize}
We evaluate each method over five experiments with different random seeds. More implementation details are summarized in Appendix~\ref{sec:implementation}.

\subsection{Spurious Shifts}
\label{sec:exp_spurious}

We first evaluate the robustness of the forecasting model trained by our invariant loss under different ranges of spurious shifts. In particular, we compare our method against the two following baselines:
\begin{itemize}[noitemsep,topsep=0pt,leftmargin=14pt]
    \item \textit{Vanilla ERM}: the conventional learning method that minimizes the average prediction error on all training samples (Eq.~\ref{eq:erm});
    \item \textit{Counterfactual Analysis} \cite{chen_human_2021}: a causality-inspired trajectory forecasting method that estimates and subtracts biased features through counterfactual interventions.
\end{itemize}
For a fair comparison with the recent counterfactual approach \cite{chen_human_2021}, we implement our method based on the same open-sourced code.
Specifically, we follow their choice of the base model, \textit{i.e.}, STGAT \cite{huang_stgat_2019}, in our experiments.
The encoder of the STGAT contains two LSTMs and one graph attention network (GAT) to account for the historical trajectory and social interaction clues, while the decoder is modeled by an LSTM to rollout the future trajectory.

\begin{figure}[t]
    \centering
        \includegraphics[height=4.4cm]{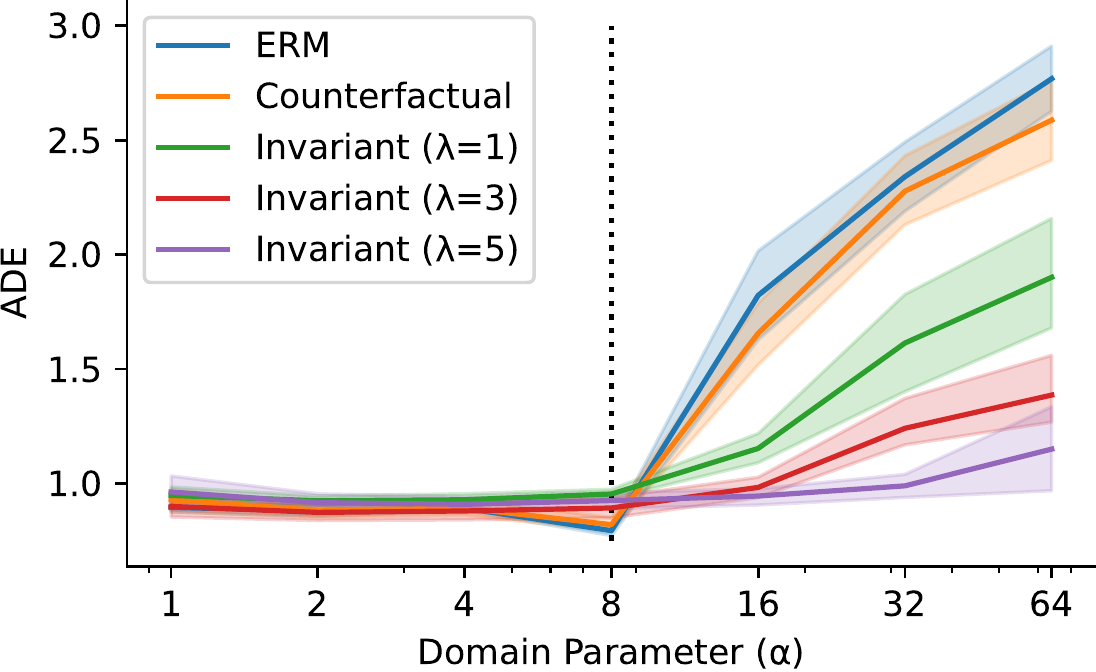}
    \caption{\vspace{-2pt}Comparison of different methods on the ETH-UCY dataset with controlled spurious correlations. 
    Our invariant learning approach substantially outperforms the conventional ERM and the counterfactual approach \cite{chen_human_2021} in the out-of-distribution regime $\alpha \in (8,64]$, while being on par within the training domains.}
    \label{fig:domain_shift}
\end{figure}


\paragraph{Setup.}
The original ETH-UCY dataset contains five subsets collected at different locations \cite{pellegrini_improving_2010,lerner_crowds_2007}. While recent work \cite{chen_human_2021} has highlighted the intrinsic differences between these subsets, it is still not trivial to pinpoint the detailed biases in each environment. 
To clearly examine the robustness of a motion forecasting model against non-causal biases, we modified the ETH-UCY dataset by introducing a third input variable measuring the level of observation noises, given that variations of spurious noise often occur in real-world problems \cite{arjovsky_invariant_2020}. 
Specifically, at each time step $t$, we simulate the observation uncertainty $\bm \sigma_t$ as a linear function of the local trajectory curvature (more details in \cref{sec:implementation}):
\begin{equation}
\begin{split}
    \bm \gamma_t &:= (\dot{x}_{t+\delta t}-\dot{x}_{t})^2+(\dot{y}_{t+\delta t}-\dot{y}_{t})^2, \\
    \bm \sigma_t &:= \bm \alpha \cdot (\bm \gamma_t +1),    
\end{split}
\label{eq:spurious}
\end{equation}
where $\dot{x}_{t}=x_{t+1}-x_{t}$ and $\dot{y}_{t}=y_{t+1}-y_{t}$ reflect the velocity of the agent within the temporal window of length $\delta t=8$, and $\bm \alpha$ is a domain specific parameter to control the strength of spurious features. 
We train the model in four environments (`hotel',`univ',`zara1' and `zara2') with $\alpha \in \{1,2,4,8\}$ and test it on the remaining one (`eth') with $\alpha \in \{1,2,4,8,16,32,64\}$. 

\paragraph{Results.}
In Figure \ref{fig:domain_shift} we show the prediction accuracy on the test sets resulting from different learning methods.
All the methods perform strongly in the training domains, \textit{i.e.}, $\alpha \in [1,8]$.
However, in the out-of-distribution regime, the accuracies of the two baseline methods significantly drop with an increased value of the domain parameter.
Notably, when the strength of the spurious feature is 8 times of the maximum strength seen during training, the ADE of the vanilla ERM rises to 3.0, approximately three times worse than its performance in the training domains. 
While the counterfactual approach \cite{chen_human_2021} is slightly better than the vanilla ERM, it also suffer suffers from a large ADE at $\sim$2.5.
In comparison, the forecasting models trained by our invariant method are clearly less sensitive to the changes of the domain parameter.
It is also visually distinct that a large emphasis ($\lambda$ in Eq.~\ref{eq:irmv1}) on the invariant penalty term during training leads to a more robust model under spurious shifts.

In Figure~\ref{fig:qualitative} we visualize the qualitative results on a particular test example of the augmented ETH-UCY dataset. 
While the input trajectories remain the same across all domains, a growing strength of the spurious feature causes a dramatic shrinkage of the predicted trajectories from the baseline methods. In contrast, the outputs of our method stay almost constant under spurious shifts.

\subsection{Style Shifts}
\label{sec:exp_style}

We further evaluate the forecasting models trained by our method in the presence of style shifts. 
As elaborated in \cref{sec:method_style}, it is often impractical for the model to directly generalize to new styles.
We therefore consider two different scenarios: robustness in the zero-shot and transfer learning results in the low-shot setting.

\paragraph{Setup.} The motion styles of existing real-world data are often largely unknown. We thus create some synthetic trajectories using ORCA \cite{van_den_berg_reciprocal_2011}, a popular multi-agent simulator, in circle-crossing scenarios \cite{chen_crowd-robot_2019} with varied style parameters.
Specifically, we consider three training styles where the simulated agents keep different minimum separation distances from each other, \textit{i.e.}, \{0.1, 0.3, 0.5\} meters. 
For each training domain, we generate 10,000 trajectories for training, 3,000 trajectories for validation and 5,000 trajectories for test. 
We evaluate each model on the training environments (IID) as well as the two new test environments with the minimum separation distance of 0.4 (OOD-Inter) and 0.6 (OOD-Extra).
We use a variant of PECNet \cite{mangalam_it_2020} as our base model which employs a MLP as the basic building block for the  encoders and decoder in our modular design. 
More implementation details are reported in Appendix~\ref{sec:implementation}.

\subsubsection{Out-of-distribution Generalization}
\label{sec:exp_style_generalization}

\paragraph{Results.} 
In Table \ref{table:synthetic} we report the results of different forecasting models in both the training domain and the out-of-distribution regimes. 
Similar to \cref{sec:exp_spurious}, the vanilla baseline suffers from much larger prediction errors in the OOD test sets than in the training ones. 
Given style changes, our invariant method alone does not yield clear advantages neither, as it tends to ignore the domain-specific style confounder. 
In contrast, our modular architecture design allows the model to effectively incorporate the domain-specific style features and thus achieves superior performance in all environments.
In particular, training the first encoder $\Encinv$ in our modular network with the invariant loss results in the best robustness in the OOD regime while being competitive in the training domains.
It is also evident that there remains a clear performance gap between the IID and OOD-Extra domain, which suggests the importance of building adaptive models investigated next.

\begin{table}[t]
    \small
    \centering

\resizebox{\scalefactor\linewidth}{!} 
{
    \begin{tabular}{cccc}
    \toprule
        Method & IID & OOD-Inter & OOD-Extra \\        
    \midrule
        Vanilla (ERM) & 0.113 $\pm$ 0.004 & 0.112 $\pm$ 0.003 & 0.192 $\pm$ 0.013 \\
        \rowcolor{gray!10}
        Invariant (ours) & 0.115	$\pm$ 0.005 & 0.114 $\pm$ 0.004 & 0.191 $\pm$ 0.007 \\
        Modular (ours) & \textbf{0.063 $\pm$	0.005} & 0.070 $\pm$	0.006 & 0.112 $\pm$ 0.004 \\
        \rowcolor{gray!10}
        Inv + Mod (ours) & 0.065	$\pm$ 0.007 & \textbf{0.069	$\pm$ 0.007} & \textbf{0.107 $\pm$	0.007} \\        
    \bottomrule
    \end{tabular}
}
    \caption{Quantitative comparison of different methods under style shifts. 
    Models are evaluated by ADE (lower is better) over 5 seeds. 
    Both the vanilla baseline and our invariant approach alone suffer from large errors, since they either \textit{average} the domain-varying styles or \textit{ignore} them. 
    Our modular network incorporates \textit{distinctive} style features into prediction and hence yields much better results.
    In particular, enforcing the causal invariance of the first encoder $\Encsty$ leads to the best OOD robustness, while being
    highly competitive in the training environments.
    }
    \label{table:synthetic}
\end{table}

\begin{figure}[t]
    \centering
    \vspace{8pt}
    \includegraphics[width=\linewidth*\real{0.8}*\real{\scalefactor}]{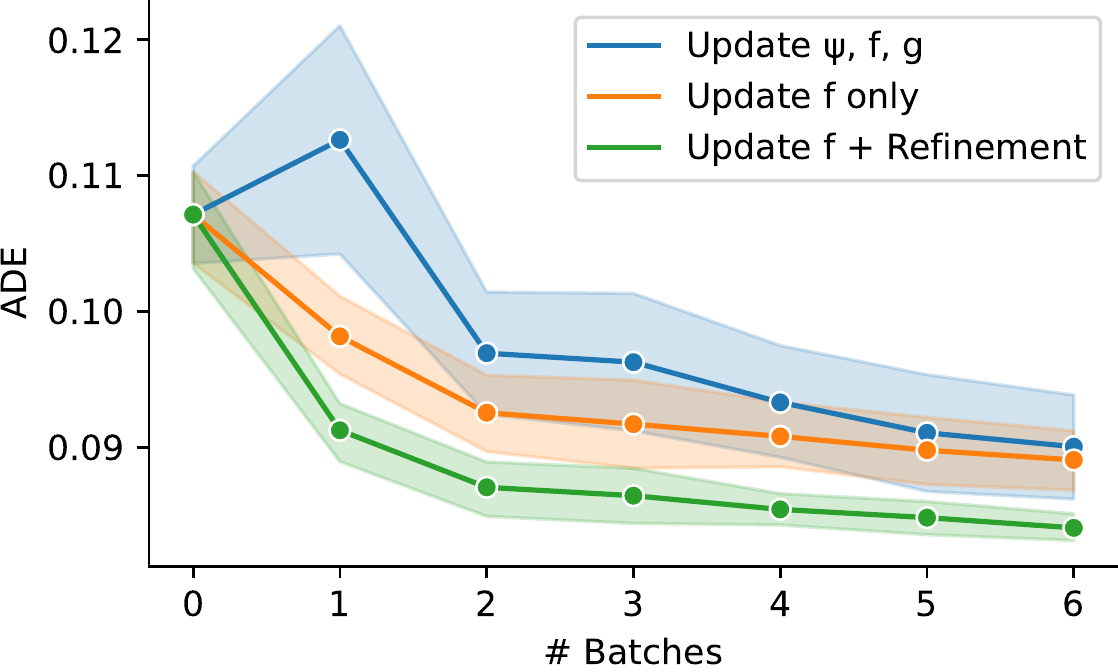}
    \caption{
    Quantitative results of different methods for transfer learning to a new motion style, given limited batch of samples.
    Our modular adaptation strategy (updating the style modulator $\Modsty$) yields higher sample efficiency than the conventional counterpart in the low-data regime.
    Moreover, refining the predicted output for 3 iterations further reduces the prediction error on the fly. 
    }
    \label{fig:transfer}
\end{figure}

\subsubsection{Low-shot Transfer}
\label{sec:exp_transfer}

As shown above (\cref{sec:exp_style_generalization}), it is practically unrealistic for a forecasting model to directly generalize to all kinds of distribution shifts.
We next evaluate the effectiveness of our proposed adaptation method in the context of low-shot transfer.
We again consider the challenging OOD-Extra style shift scenario and compare the following options:
\begin{enumerate}[noitemsep,topsep=0pt,leftmargin=24pt,label=(\alph*)]
    \item conventional approach fine-tuning all parameters;
    \item our modular adaptation strategy fine-tuning $\Modsty$ only; \label{itm:modulator}
    \item our test-time refinement on top of our method \ref{itm:modulator}. 
\end{enumerate}
We evaluate all methods given a limited number of samples, \textit{i.e.}, $\{1, 2, \dots, 6\} \times \mathrm{BS}$, where $\mathrm{BS} = 64$ is the batch size.

\paragraph{Effect of modular adaptation.}

Figure~\ref{fig:transfer} shows the results of different adaptation methods in the low-shot setting.
In the case of only one batch of observations, fine-tuning all style-related parameters ($\Encsty, \Modsty$ and $\Dec$) leads to noisy outcomes and worsens the results on average.
In comparison, updating $\Modsty$ while keeping the remaining majority of the parameters fixed yields clearly better performance in the low-data regime. 
For instance, fine-tuning $\Modsty$ on two batches of the new style achieves the same level of prediction accuracy as fine-tuning the whole model on five batches.  



\paragraph{Effect of test-time refinement.}

Finally, we evaluate the effectiveness of test-time refinement based on the style contrastive loss. As shown in Figure~\ref{fig:transfer}, our refinement techniques leads to substantial error reductions on top of the fine-tuned models. Figure~\ref{fig:refinement} shows the qualitative effect on a two-agent scenario, where the predicted trajectories gradually get closer to the ground truth based on a scene observation of the target style as a reference.
This result suggests a strong promise of reusing the structural knowledge learned in our modular forecasting model at test-time.

\paragraph{Additional results and discussions.}

Please refer to Appendix for additional experiments, implementation details as well as discussions about limitations and future work.

\begin{figure}[t]
\vspace{-4pt}
    \centering
    \includegraphics[width=\scalefactor\linewidth]{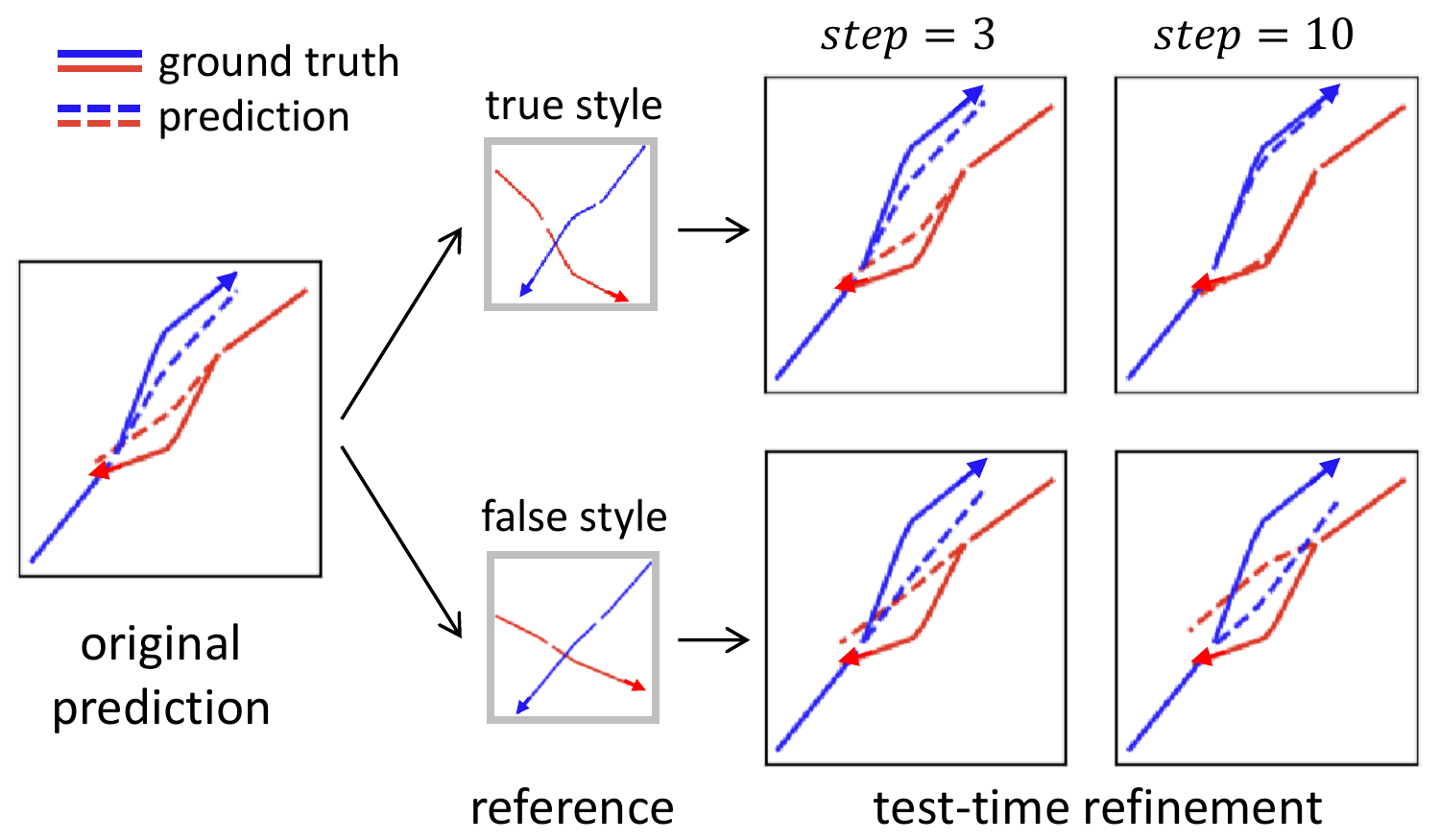}
    \caption{Qualitative effects of test-time refinement in a two-agent scenario. 
    The initial predicted output suffers from a clear prediction error.
    Given a scene observation of the true style (large separation distance) as a reference, our method gradually closes the discrepancy between the predicted trajectory and the ground truth.
    Conversely, when conditioned on a scene of a different style (small separation distance) as the reference, our method manages to steer the output towards the corresponding false style as well.
    }
    \label{fig:refinement}
\end{figure}

\section{Conclusions}
\label{sec:conclusion}
We present a causality-inspired learning method for motion forecasting. 
Given data collected from multiple locations, our invariant loss yields stronger generalization than the previous statistical and counterfactual methods in the presence of spurious distribution shifts. 
In addition, our modular architecture design coupled with the proposed style contrastive loss enhances the robustness and transferability of learned motion representations under style shifts.
Our results suggest that incorporating causal invariance and structure into representation learning is a promising direction towards robust and adaptive motion forecasting.

\paragraph{Acknowledgments.}
\label{sec:ackn}
This work is supported by the Swiss National Science Foundation under the Grant 2OOO21-L92326.
We thank Bastien Van Delft, Brian Alan Sifringer and Yifan Sun for thoughtful feedback on early drafts, Parth Kothari and Hossein Bahari for valuable suggestions on experiments, as well as reviewers for insightful comments.


\clearpage

\appendix

\section{Additional Experiments} \label{sec:ablation}

\revision{

\begin{table}[t]
    \small
    \centering
    \begin{tabular}{c|ccc}
    \toprule
    Method & Vanilla (ERM) & Counterfactual & Invariant (ours) \\
    \midrule
    ADE ($\downarrow$) & 0.536 $\pm$ 0.015 & 0.512 $\pm$ 0.057 & \textbf{0.457} $\pm$ 0.054 \\
    \midrule
    FDE ($\downarrow$) & 1.088 $\pm$ 0.039 & 1.029 $\pm$ 0.136 & \textbf{0.918} $\pm$ 0.098 \\
    \bottomrule
    \end{tabular}
    \caption{Comparison of different methods on the original ETH-UCY dataset. The STGAT \cite{huang_stgat_2019} trained by our invariant approach substantially outperforms the vanilla ERM and the counterfactual counterparts \cite{chen_human_2021}. The number of sampled trajectories is set to 1 for computational efficiency. Results are averaged over 5 seeds.}
    \label{tab:ori_eth_ucy}
    \vspace{6pt}
\end{table}

\begin{figure}[t]
    \centering
    \includegraphics[width=\linewidth*\real{0.6}*\real{\scalefactor}]{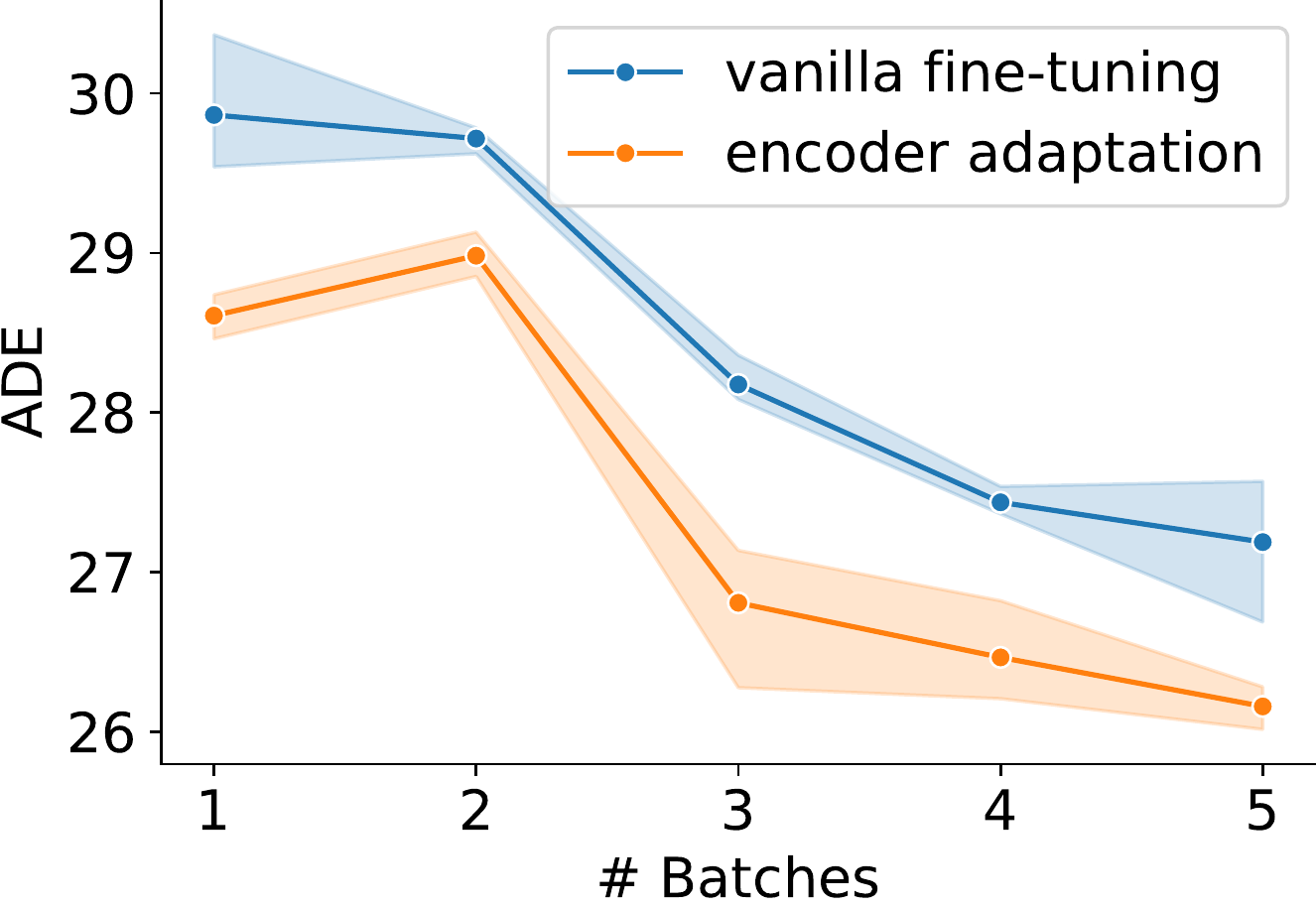}
    \caption{Quantitative results of low-shot transfer on the SDD \cite{robicquet_learning_2016} dataset.
    Our modular adaptation strategy yields higher sample efficiency than the conventional fine-tuning.
    }
    \label{fig:sdd}
\end{figure}

\paragraph{Robustness on the original ETH-UCY dataset.}

In addition to the robustness experiments under controlled distribution shifts of spurious features (\cref{sec:exp_spurious}) or style features (\cref{sec:exp_style}), we also evaluate our method on the original ETH-UCY dataset, where each subset is subject to some unknown selection biases.
We train the STGAT \cite{huang_stgat_2019} on four subsets (`eth',`univ',`zara1' and `zara2') and test it on the rest (`hotel'). The results in Table~\ref{tab:ori_eth_ucy} confirms the strength of our method on real-world data.

\paragraph{Low-shot transfer on the SDD dataset.}

Apart from simulated style shifts in \cref{sec:exp_style}, we further evaluate our method on the SDD dataset under substantial style shifts. We create four different domains according to the agent type and average speed. We use three domains for training and the last one for evaluation. We apply our method on top of the Y-Net \cite{mangalam_goals_2021}, and compare our modular adaptation strategy against the standard fine-tuning of the entire model for low-shot transfer. The results in Fig.~\ref{fig:sdd} demonstrate the scalability of our method to real-world style shifts.

}

\paragraph{Larger style shifts.}

As a supplement to Table~\ref{table:synthetic}, we summarize the detailed results under larger style shifts in Table~\ref{tab:larger_style}.
Among these OOD test domains ($d  > 0.5$), the farther the style parameter is from the training ones, the larger improvement we obtain from using the full version of our method. 
This result confirms the advantage of our modular design with an enforced structure of the invariant and style knowledge for robust generalization.

\paragraph{Style contrastive pre-training.}

As described in \cref{sec:method_style}, one advantage of incorporating the proposed style contrastive loss is to ease the training of our modular model that consists of multiple sub-networks. 
In Figure~\ref{fig:contrastive} we compare the performance of the models during training with and without the style contrastive pre-training.
The model pre-trained on the style contrastive task learns significantly faster than the counterpart without it.

\begin{table}[t]
    \small
    \centering
    \resizebox{\scalefactor\linewidth}{!}{
    \begin{tabular}{c|ccc}
    \toprule
        Method & $d =$ 0.6 & $d =$ 0.7 & $d =$ 0.8 \\        
    \midrule
        Vanilla (ERM) & 0.192 $\pm$ 0.013 & 0.246 $\pm$ 0.020 & 0.309 $\pm$ 0.025 \\
        \rowcolor{gray!10}
        Invariant (ours) & 0.191 $\pm$ 0.007 & 0.245 $\pm$ 0.009 & 0.309 $\pm$ 0.011 \\
        Modular (ours) & 0.112 $\pm$ 0.004 & 0.169 $\pm$ 0.011 & 0.242 $\pm$ 0.020 \\
        \rowcolor{gray!10}
        Inv + Mod (ours)  & \textbf{0.107} $\pm$ 0.007 & \textbf{0.156} $\pm$ 0.013 & \textbf{0.221} $\pm$ 0.020 \\
    \bottomrule
    \end{tabular}
    }
    \caption{ADE scores of different methods on OOD-Extra domains. 
    The full version (invariant + modular) of our method yields more performance gains with an increasing degree of style shifts.
    }
    \label{tab:larger_style}
\end{table}

\begin{figure}[t]
    \vspace{6pt}
    \centering
    \begin{subfigure}[b]{0.495\linewidth}
        \centering
        \includegraphics[width=\widthfactor\textwidth]{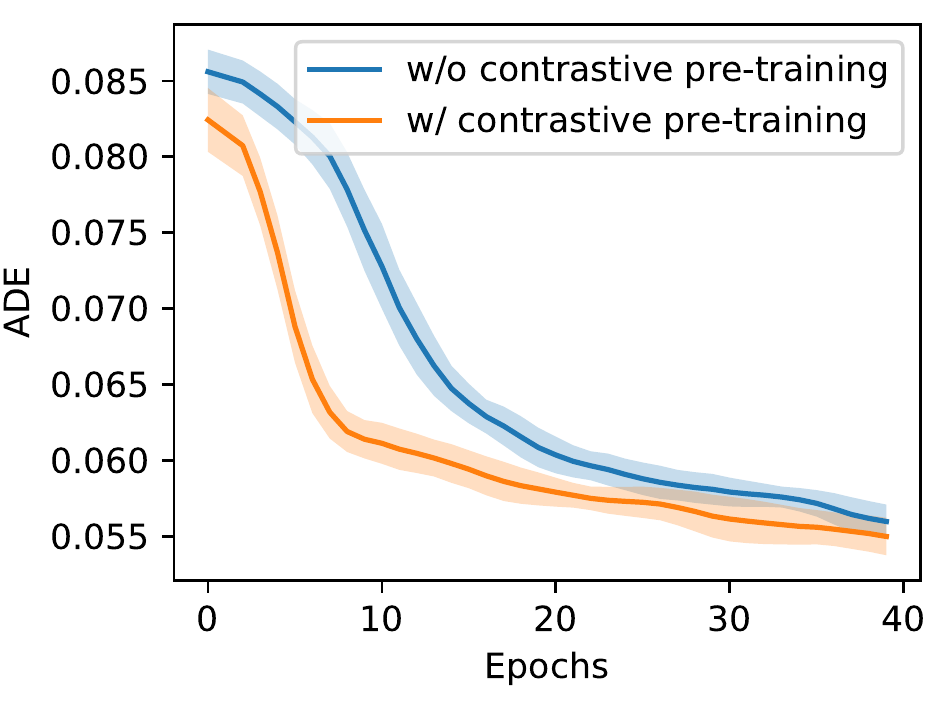}
        \caption{\vspace{1em}IID style $d=0.1$}
    \end{subfigure}
    \begin{subfigure}[b]{0.495\linewidth}
        \centering
        \includegraphics[width=\widthfactor\textwidth]{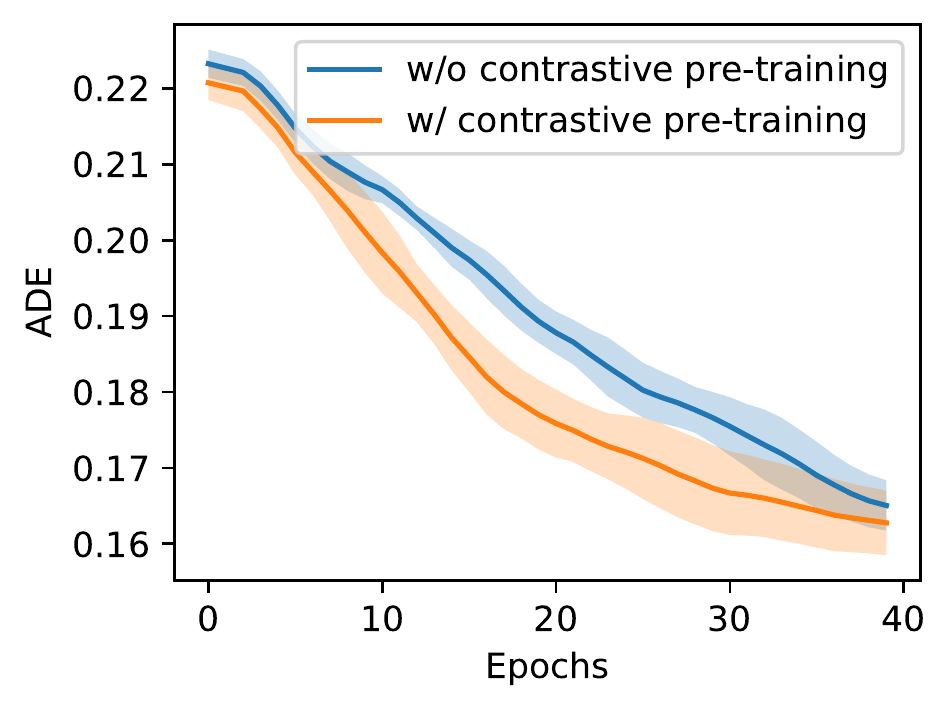}
        \caption{\vspace{1em}OOD style $d=0.7$}
    \end{subfigure}
    \caption{Comparison of the models with and without style contrastive pre-training. The model pre-trained on the style contrastive task converges faster than the counterpart during the end-to-end training in both domains.}
    \label{fig:contrastive}
\end{figure}

\section{Experimental Details} \label{sec:implementation}

\subsection{Spurious Shift Experiments}
\label{sec:impl_spurious}

\revision{

\begin{figure*}[t]
\vspace{\figmargin}
\centering
\resizebox{.95\textwidth}{!}{
\includegraphics{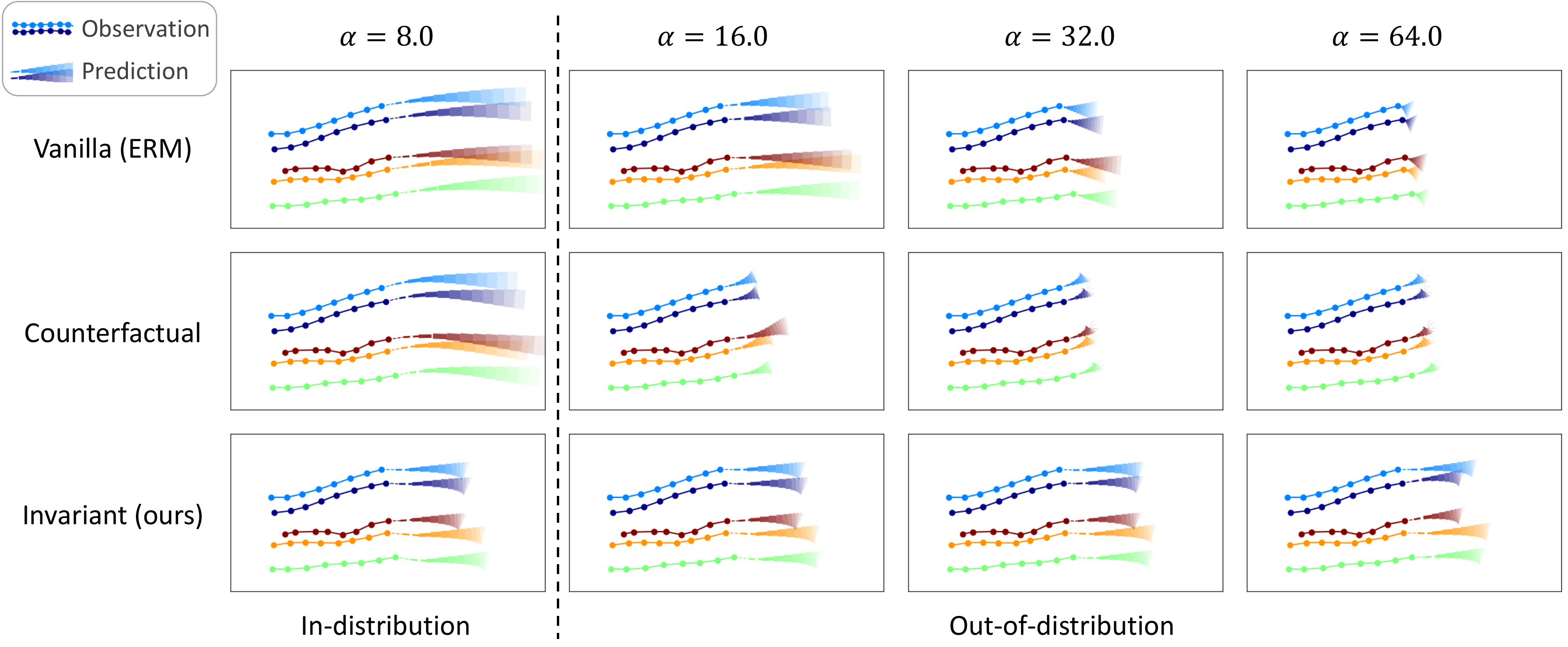}
}
\caption{
Visualization of the predicted trajectories from different methods in a particular test case of the ETH-UCY dataset with controlled spurious features.
Despite the \textit{same} past trajectory observation and ground truth future, the predicted trajectories from the two baselines abruptly slow down, when the strength of the spurious feature at test time is larger than that in the training domains. 
In comparison, our invariant learning approach results in much more robust solutions, even under substantial spurious shifts (\textit{e.g.}, $\alpha=64.0$).
}
\label{fig:qualitative}
\end{figure*}
    
\paragraph{Experimental design.}

To clearly examine the robustness of motion forecasting models against spurious shifts (\cref{sec:exp_spurious}), we introduce an additional input variable $\sigma_t$, concatenated to the 2D coordinates. Its value is defined as a linear function of the trajectory curvature $\gamma_t$. By changing the scaling coefficient $\alpha$, we artificially control a varied degree of spurious shifts.
This controlled setting allows us to simulate the spurious correlation arising from two co-occurring phenomena in crowded spaces: observations become noisy due to occlusions; trajectories become non-linear because of interactions.
Given this coincidence, statistical models may exploit the level of noise $\sigma_t$ to ease predictions. Yet, such non-causal models are brittle. Any changes of the noise pattern (\textit{e.g.}, due to perception algorithm updates illustrated in Figure~\ref{fig:illustration}) may degrade forecasting accuracy.
}

\paragraph{Architecture.}
For the experiments reported in \cref{sec:exp_spurious}, we use the standard STGAT \cite{huang_stgat_2019} architecture for a fair comparison with the counterfactual analysis approach \cite{chen_human_2021}.
In order to take the $x$ and $y$ coordinate as well as the observational uncertainty $\sigma_t$ as inputs, we adjust the input dimension of the first LSTM module to three.
All the remaining configurations align with the original STGAT. 
Following the previous work \cite{chen_human_2021}, we train the model in three steps: (i) pre-train the first LSTM, (ii) pre-train the GAT together with the second LSTM, (iii) train the whole model.

\paragraph{Hyper-parameters.}
We use the same hyper-parameters as in the original STGAT \cite{huang_stgat_2019}.
For the invariant penalty coefficient $\lambda$, we run grid search in a range from 0.001 to 100.
Since the focus of our experiments is on the robustness and adaptability under distribution shifts rather than the performance on training domains, we only predict one trajectory output per instance instead of multiple ones \cite{gupta_social_2018} during training, which reduces computational expenses for the comparison of different methods.
Other detailed hyper-parameters are summarized in Table~\ref{tab:hyper_spurious}.

\subsection{Style Shift Experiments}
\label{sec:impl_style}

\paragraph{Architecture.} For the experiments in \cref{sec:exp_style}, we use a PECNet-like \cite{mangalam_it_2020} feedforward network as our base model. Specifically, we model all components using MLPs.
We train the modular network in four detailed steps: 
(i) train the invariant encoder together with the decoder,  
(ii) subsequently pre-train the style encoder and the projection head, (iii) followed by the style modulator, and (iv) finally train the entire model end-to-end.

\paragraph{Hyper-parameters.} We keep most hyper-parameters identical to the setup in \cref{sec:impl_spurious}.
We further tune the learning rates for each module separately due to their distinct properties. 
Detailed settings for training, adaptation and refinement are summarized in Table~\ref{tab:hyper_style}.

\subsection{Other Details} 
We train all of our models on a single NVIDIA Tesla V100 GPU. Each run takes around one hour. 
The source code of our method as well as baselines can be found at \url{https://github.com/vita-epfl/causalmotion} and \url{https://github.com/sherwinbahmani/ynet_adaptive}.

\begin{table}[t]
    \small
    \centering
    \begin{tabular}{wl{4.8cm}|wl{2.2cm}}
        config & value \\        
    \specialrule{.1em}{.05em}{.05em}  
        batch size & 64 \\
        epochs per stage & 150, 100, 150 \\
        learning rate & 0.001 \\
    \end{tabular}
    \caption{Hyper-parameters in spurious shift experiments.
    }
    \label{tab:hyper_spurious}
\end{table}

\begin{table}[t]
    \small
    \centering
    \begin{tabular}{wl{4.8cm}|wl{2.2cm}}
        config & value \\        
    \specialrule{.1em}{.05em}{.05em}  
        batch size & 64 \\
        epochs per stage & 100, 50, 20, 300 \\
        contrastive loss coefficient & 1.0 \\
        learning rate baseline & 0.001 \\
        learning rate style encoder  & 0.0005 \\
        learning rate projection head  & 0.01 \\
        learning rate style modulator (train) & 0.01 \\
        learning rate style modulator (adapt) & 0.001 \\
    \end{tabular}
    \caption{Hyper-parameters in style shift experiments.
    }
    \label{tab:hyper_style}
\end{table}

\section{Additional Discussions} \label{sec:discussion}

To the best of our knowledge, our work provides the first attempt to incorporate causal invariance and structure into the design and learning of motion forecasting models.
Despite encouraging results, our work is still subject to a couple of limitations.

\paragraph{Limitations \& future work.}
One major technical limitation lies in the granularity of the considered causal representations.
While our method places great emphasis on three prominent groups of high-level latent features, we have largely overlooked the structure of fine-grained features. 
One interesting direction for future work is to further exploit detailed causal structure for motion forecasting, for instance, (i) disentangling the left or right-hand traffic rules from social distance conventions within the group of style confounders, (ii) encouraging sparse interplay between sub-modules, \textit{e.g.}, pruning the connections between inertia features and left or right-hand traffic rules, given their presumably minute significance. 

Another limitation of our work is tied to the scale and diversity of experiments.
Thus far, we have demonstrated the strengths of our method on two human motion datasets and two base models as proofs of concept.
Nevertheless, our method is highly generic and we hypothesize that it can also bring similar benefits to other types of motion problems and datasets, \textit{e.g.}, vehicles \cite{caesar_nuscenes_2020}, sports \cite{yue_learning_2014} and driving simulations \cite{mcduff_causalcity_2021}. 
Extending the current empirical findings to more contexts can be another valuable avenue for future work.

\paragraph{Societal impact.}

Out-of-distribution robustness remains a salient weakness of motion forecasting models while having a crucial impact on the safety of autonomous systems, especially in the context of autonomous driving. Even though these machines operate accurately in their training environments, deploying them in unseen test conditions can result in undesired behavior, which may ultimately lead to fatal consequences in specific scenarios. With our work, we contribute to reducing this performance gap. However, we are aware of the remaining deficits of our motion forecasting approach in significantly changing conditions that should not be neglected when utilizing such systems in real-world applications.

{\small
\bibliographystyle{ieee_fullname}
\bibliography{yuejiang,sherwin}
}

\end{document}